\documentclass[11pt]{article}

\usepackage[preprint]{acl}

\usepackage{times}
\usepackage{latexsym}

\usepackage[T1]{fontenc}

\usepackage[utf8]{inputenc}

\usepackage{microtype}

\usepackage{inconsolata}

\usepackage{graphicx}

\usepackage{hyperref}
\hypersetup{colorlinks=true, citecolor=mycitecolor, pdfborder={0 0 0}}
\usepackage{url}

\usepackage{amsfonts}
\usepackage{nicefrac}
\usepackage{xcolor,colortbl}
\usepackage{arydshln}
\usepackage{color}
\usepackage[normalem]{ulem}
\usepackage{amsmath}
\usepackage{amssymb}
\usepackage{epsfig}
\usepackage{pifont}
\usepackage{multirow,multicol,makecell}
\usepackage{bbm}
\usepackage{mathtools}
\usepackage{diagbox}
\usepackage{enumitem}
\usepackage{soul}
\usepackage{tikz}
\usepackage{bbding}
\usepackage{wrapfig}
\usepackage{colortbl}

\usepackage{xspace}
\usepackage{hhline}
\usepackage{caption}
\usepackage{wrapfig}
\usepackage{booktabs}
\usepackage[table]{xcolor}

\usepackage{verbatim}
\usepackage{listings}

\definecolor{tablerowcolor}{RGB}{235,245,252}
\definecolor{tablerowcolor1}{RGB}{235,248,235}
\definecolor{tablerowcolor2}{RGB}{252,235,240}

\definecolor{mygray}{gray}{0.9}
\definecolor{mytableblue}{rgb}{0.83, 0.90, 0.94}
\definecolor{mytablegreen}{rgb}{0.90, 0.97, 0.87}
\definecolor{mygreen}{RGB}{84,130,53}
\definecolor{mycitecolor}{RGB}{0,101,177}

\newcommand*{\affmark}[1][*]{\textsuperscript{#1}}

\newcommand{\eg}{\textit{e}.\textit{g}.}

\newcommand{\etc}{\textit{etc}}

\title{From Pixels to Words -- Towards Native One-Vision Models at Scale}

\author{
\centerline{
Haiwen Diao\textsuperscript{\rm 1,2}\thanks{Work was done during Haiwen's remote collaboration with SenseTime Research. \affmark[\dag]Corresponding author.},\;
Jiahao Wang\textsuperscript{\rm 2},\;
Penghao Wu\textsuperscript{\rm 1,2},\;
Yuhao Dong\textsuperscript{\rm 1}
}\\
\centerline{
\textbf{
Yuwei Niu\textsuperscript{\rm 2},\;
Yue Zhu\textsuperscript{\rm 2},\;
Zhongang Cai\textsuperscript{\rm 2},\;
Weichen Fan\textsuperscript{\rm 1,2},\;
Linjun Dai\textsuperscript{\rm 2}
}
}\\
\centerline{
\textbf{
Silei Wu\textsuperscript{\rm 2},\;
Xuanyu Zheng\textsuperscript{\rm 2},\;
Mingxuan Li\textsuperscript{\rm 2},\;
Yuanhan Zhang\textsuperscript{\rm 1},\;
Bo Li\textsuperscript{\rm 1},\;
Hanming Deng\textsuperscript{\rm 2}
}
} \\
\centerline{
\textbf{
Huchuan Lu\textsuperscript{\rm 3},\;
Quan Wang\textsuperscript{\rm 2},\;
Lei Yang\textsuperscript{\rm 2},\;
Lewei Lu\textsuperscript{\rm 2},\;
Dahua Lin\textsuperscript{\rm 2},\;
Ziwei Liu\textsuperscript{\rm 1}\affmark[\dag]
}
}\\
\centerline{
\textsuperscript{\rm 1}S-Lab, NTU \,
\textsuperscript{\rm 2}SenseTime Research \,
\textsuperscript{\rm 3}DLUT} \\
\parbox{\textwidth}{
\centering
\begin{tabular}{ll}
\raisebox{-0.15em}{\includegraphics[height=1.05em]{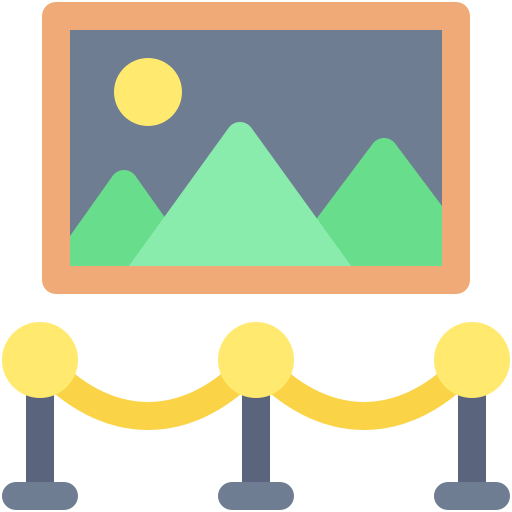}} \textbf{Website:} & \url{https://github.com/EvolvingLMMs-Lab/NEO}
\end{tabular}
}
}

\begin{document}
\maketitle
\begin{abstract}

Current vision–language models (VLMs) typically stitch together separate image encoders and language decoders via multi-stage alignment, a modular framework that inevitably fragments pixel-level signals across frames and scatters early pixel–word interactions.
In parallel, native VLMs, despite impressive performance on single images, remain largely unexplored in multi-image, video understanding, and spatial intelligence.
Hence, we introduce \textbf{NEO-ov}, a native foundation model that learns cross-frame and pixel-word correspondence end-to-end, without any external encoders, auxiliary adapters, or post-hoc fusion. 
By eliminating module boundaries entirely, \textbf{NEO-ov} enables fine-grained and unified spatiotemporal modeling to emerge natively inside the model.
Notably, \textbf{NEO-ov} largely narrows the gap to modular counterparts while excelling at fine-grained visual perception, validating that native “one-vision” architectures are not only feasible but competitive at scale. 
Beyond empirical performance, we unveil systematic architectural analyses and detailed training recipes to facilitate subsequent native multimodal modeling.

\end{abstract}


\section{Introduction}
\label{sec:intro}

Recently, vision–language models (VLMs) have evolved from basic image perception towards advanced understanding of multi-image analysis, video understanding, and spatial intelligence.
Existing models typically adopt an encoder–decoder architecture, where pretrained image~\cite{VLP:CLIP,VLP:SigLIP} or video~\cite{VLM:videochat,VLM:LLaVAPrime} encoders produce visual representations that are subsequently processed by a projector~\cite{VLM:LLaVA-1.5,VLM:Deepstack,VLM:InstructBLIP,VLM:langbridge} and a large language model (LLM)~\cite{TransF:LLaMA2,TransF:Qwen3} for visual understanding and reasoning.

Despite strong performance, this modular design imposes inherent constraints on 
\textbf{1) Flexibility:}
vision encoders are expected to process heterogeneous inputs, from single images to image sets or videos. Yet existing designs force a false dichotomy: image encoders favor static, frame-level representations and lack spatiotemporal reasoning, while video encoders overemphasize temporal dynamics and generalize poorly to single-image or interleaved inputs. Besides, both struggle in early pixel–word interaction and unified visual understanding scenarios.
\textbf{2) Efficiency:}
decoupling vision and language modules fragments training and incurs substantial post-alignment overhead. Furthermore, extending visual encoders to long-duration or high-resolution inputs remains prohibitively expensive for streaming and proactive video understanding, as KV caching is not applicable.
\textbf{3) Scalability:}
modularity entangles scaling, optimization, and deployment by requiring delicate capacity balancing between VEs and LLMs. These frictions fundamentally preclude structural simplicity and deep vision–language integration, motivating a unified, monolithic backbone.

To address them, native VLMs have recently emerged as a compelling alternative. Early exemplars, \eg, Fuyu~\cite{VLM:Fuyu-8b} and EVE~\cite{VLM:EVE} demonstrate that visual and textual inputs can be jointly modeled within one single and monolithic framework without explicit vision encoders. Building on this paradigm, subsequent efforts learn visual representations from scratch while mitigating vision–linguistic interference through visual feature distillation~\cite{VLM:EVE,VLM:BREEN,VLM:VoRA}, modality-agnostic embeddings~\cite{VLM:NEO,VLM:HoVLE,VLM:HaploVL} and modality-specific decomposition~\cite{VLM:EVEv2,VLM:Mono-InternVL,VLM:Mono-InternVL-1.5}.
Notably, recent studies~\cite{VLM:Video-Panda,VLM:ELVA} extend native VLMs to video domains, enabling end-to-end modeling of fine-grained video–language interactions and temporal dependencies. However, these approaches remain constrained by distillation from static visual encoders, inheriting strong inductive biases rooted in pretrained image semantics. 
More importantly, unifying single-image, multiple-image, video understanding, and spatial intelligence simultaneously remains an open frontier for native VLMs toward truly unified one-vision foundation models across diverse multimodal applications.

Hence, we introduce \textbf{NEO-ov}, a native vision-language foundation model that eliminates pretrained encoders and unifies spatial and temporal modeling within a single monolithic backbone. 
Built on multiple native primitives, \textbf{NEO-ov} jointly learns visual perception, temporal dynamics, and cross-modal alignment directly from raw inputs through end-to-end training.
Despite being fully encoder-free, \textbf{NEO-ov} surpasses existing native VLMs and approaches encoder-based competitors of the same LLMs across diverse benchmarks. 
Notably, it exhibits strong spatial intelligence across both low-level geometric perception and high-level spatiotemporal reasoning, enabling robust understanding of structure, motion, and long-range visual dependencies in a unified representation space.
Together, these results suggest that multimodal intelligence may emerge not only from specialized components, but from architectures that are native, unified, and intrinsically multimodal.


\section{Related Work}
\subsection{Modular Vision-Language Models}
Existing vision-language models (VLMs) largely follow a modular design that connects external visual encoders to large language models (LLMs) through lightweight adapters~\cite{VLP:Flamingo,VLM:InstructBLIP}. Notably, LLaVA~\cite{VLM:LLaVA,VLM:LLaVA-NeXT} standardizes this paradigm via the simple \textit{Encoder-MLP-LLM} pipeline and visual instruction tuning, which is subsequently adopted by models such as InternVL series~\cite{VLM:InternVL-2.5,VLM:InternVL-3,VLM:InternVL-3.5}, Qwen-VL series~\cite{VLM:Qwen2-VL,VLM:Qwen2.5-VL,VLM:Qwen3VL}, and \etc. 
They further extend this paradigm to unified visual understanding across single-image, multi-image, and video tasks.

Despite empirical success, they remain fundamentally constrained by the encode-then-project paradigm, where visual signals are compressed before reasoning begins. Pretrained vision encoders such as CLIP~\cite{VLP:CLIP} or SigLIP~\cite{VLP:SigLIP,VLP:SigLIP-2} are optimized primarily for image–text alignment, emphasizing high-level semantics while discarding texture, local geometry, and fine spatial structure. Consequently, language models reason over semantically filtered representations rather than native visual signals, limiting fine-grained perception and precise geometric reasoning. This limitation becomes particularly pronounced in spatial intelligence settings, where cross-view and cross-frame interactions are mediated through compressed semantic features instead of native spatial correspondences, hindering the modeling of positional relations, local motion, and pixel-level consistency across space and time.

\subsection{Native Vision-Language Models}

Native multimodal models move beyond modular pipelines by learning directly from pixels and words within a unified backbone. Early works such as Fuyu~\cite{VLM:Fuyu-8b} and EVE~\cite{VLM:EVE,VLM:EVEv2} demonstrate that image patches can be integrated directly into decoder-only Transformers without separate visual encoders, establishing the feasibility of fully native multimodal modeling. Subsequent efforts further improve this paradigm through visual encoder distillation~\cite{VLM:EVE,VLM:BREEN,VLM:VoRA}, modality-specific parameterization~\cite{VLM:EVEv2,VLM:Mono-InternVL,VLM:Mono-InternVL-1.5}, and shared multimodal representations~\cite{VLM:NEO,VLM:HoVLE,VLM:HaploVL}.
Notably, NEO~\cite{VLM:NEO} further formalizes native multimodal learning and substantially narrows the gap to strong modular VLMs through shared pixel–word representations and unified cross-modal reasoning.

\begin{figure*}[t]
\centering 
\includegraphics[width=0.99\linewidth,trim= 0 0 0 0,clip]{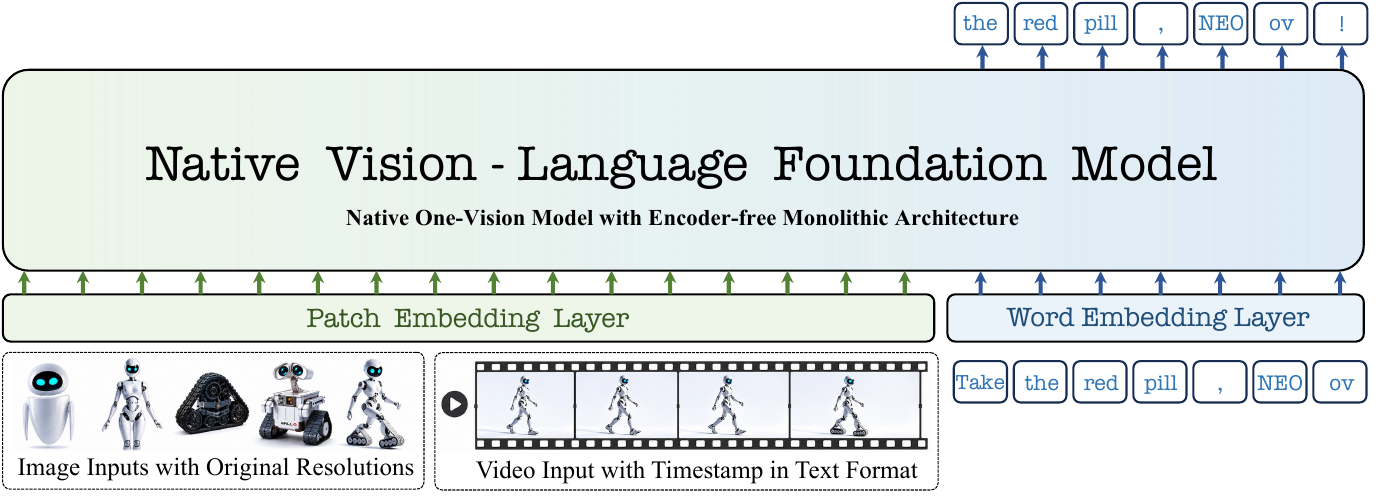} 
\caption{Overview of the NEO-ov model. 
Image or video inputs and text are encoded into token sequences via lightweight patch and word embeddings, then processed within a single decoder-only backbone composed of stacked native primitives, enabling efficient pixel–word and pixel–pixel alignment as well as spatial-temporal reasoning.}
\label{fig:frameworks}
\end{figure*}

Building on this direction, recent studies~\cite{VLM:Video-Panda,VLM:ELVA} extend native VLMs to the video domain, enabling end-to-end modeling of fine-grained video–language interactions and temporal dynamics. However, these efforts remain primarily focused on video understanding, without addressing broader multimodal settings involving single-image understanding, multi-image reasoning, spatial intelligence, and other unified perception tasks.
In contrast, NEO-ov further advances this direction by extending native modeling from predominantly single-image settings to a unified framework spanning single-image, multi-image, and video inputs, moving native VLMs closer to a general one-vision foundation architecture.

\section{NEO-ov: Native One-Vision Modeling}

NEO-ov is a native vision-language model that extends unified autoregressive modeling from single-image understanding to multi-image understanding, video understanding, and spatial intelligence. 
By organizing images, frames, regions, and text into a unified sequence, NEO-ov naturally supports cross-image reasoning, temporal understanding, and spatial localization.
To scale from single-image inputs to ordered visual sequences, we introduce a unified serialization scheme together with spatiotemporal attention mechanisms, enabling both high-level semantic reasoning and fine-grained spatial-temporal representation within one native backbone.

\subsection{Revisiting Native Modeling}

\begin{figure*}[t]
\centering 
\includegraphics[width=0.99\linewidth,trim= 0 0 0 0,clip]{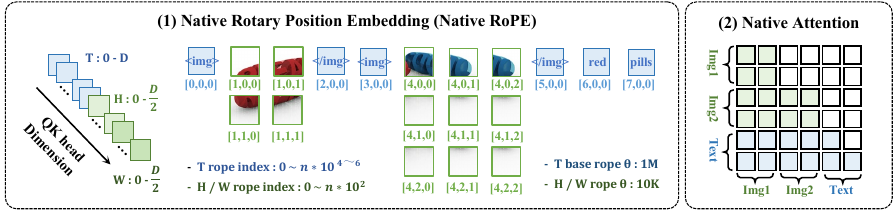} 
\caption{Overview of native rotary position embeddings and spatial-temporal attention. 
It unifies bidirectional spatial interactions within images with causal dependencies across text and video frames via $THW$-aware frequency, channel, and index allocation, enabling unified modeling across single-image, multi-image, and video understanding.}
\label{fig:primitive}
\end{figure*}

Following NEO~\cite{VLM:NEO}, NEO-ov adopts a unified native vision-language backbone.
In Figure~\ref{fig:frameworks}, we encode the image $\boldsymbol{I}$ into visual tokens by a lightweight embedding layer using two convolutional layers with a GELU activation:
\begin{equation}
\label{eq:patch_embedding_layer}
\begin{split}
\boldsymbol{x}_v &= \mathrm{Conv}_2\!\left(
\mathrm{GELU}\!\left(\mathrm{Conv}_1(\boldsymbol{I})\right)
+ \boldsymbol{\mathrm{PE}}
\right), \\
\boldsymbol{x}_t &= \mathrm{Tokenizer}(\boldsymbol{T}),
\end{split}
\end{equation}
where $\boldsymbol{x}_v \in \mathbb{R}^{n_v \times d}$, $\boldsymbol{x}_t \in \mathbb{R}^{n_t \times d}$, and $\boldsymbol{\mathrm{PE}}$ denote visual, textual, and 2D RoPE embeddings~\cite{TransF:RoPE}, respectively.
The text input $\boldsymbol{T}$ is tokenized using original LLM tokenizer. Besides, $\mathrm{Conv}_1$ extracts patches with stride 16, while $\mathrm{Conv}_2$ aggregates local features with stride 2, producing one visual token for each $32 \times 32$ image region. The visual tokens are wrapped with \texttt{<img>} and \texttt{</img>}, concatenated with the text tokens, and jointly processed by one unified backbone. 
We initialize the Pre-Buffer and Post-LLM layers from NEO~\cite{VLM:NEO} and Qwen3~\cite{TransF:Qwen3}.

For attention heads, NEO-ov still adopts an explicit $THW$-decoupled design that preserves the original LLM's head dimension as the temporal component $T$, while introducing extra head dimensions for the spatial components $H$ and $W$. This retains the temporal modeling capability inherited from the LLM while augmenting it with dedicated spatial modeling capacity. For tokens $i$ and $j$, the Query (Q) and Key (K) features are defined as:
\begin{equation}
\mathbf{q}_i=[\mathbf{q}_i^T;\mathbf{q}_i^H;\mathbf{q}_i^W],\quad
\mathbf{k}_j=[\mathbf{k}_j^T;\mathbf{k}_j^H;\mathbf{k}_j^W].
\end{equation}
Their correlation is then defined as:
\begin{equation}
s_{ij}
=
\langle \mathbf{q}_i^T,\mathbf{k}_j^T\rangle
+
\langle \mathbf{q}_i^H,\mathbf{k}_j^H\rangle
+
\langle \mathbf{q}_i^W,\mathbf{k}_j^W\rangle.
\end{equation}
The $T$ branch models textual order, cross-image relations, and cross-frame dependencies, while the $H$ and $W$ branches capture 2D spatial structure.

For rotary positional embedding (RoPE), we continue to implement Native-RoPE with separate temporal and spatial index modeling in Figure~\ref{fig:primitive} (1):
\begin{equation}
\mathrm{idx}_i = [t_i, h_i, w_i],
\end{equation}
where $t_i$ denotes the temporal or sequential positions, and $h_i, w_i$ denote the spatial coordinates. 
Text tokens retain only the temporal index, with $h_i$ = $w_i$ = $0$, whereas image tokens share the same temporal index within each image and use $h_i$ and $w_i$ to encode spatial positions. Temporal indices remain continuous across modalities, while spatial indices are independently defined within each image.

\subsection{Unified Visual Serialization}

For one single image, the model inserts one visual segment at the corresponding \texttt{<img>} position. For multi-image inputs, each \texttt{<img>} token in the prompt is replaced by an independent visual segment, following the textual order in which it appears. As a result, multiple images are represented as distinct visual units in the same sequence:
\begin{equation}
\small
\begin{split}
\mathbf{X}_{\text{multi}} = [&\,\boldsymbol{x}_{t_1},
\texttt{<img>} \,\boldsymbol{x}_{v_1}\, \texttt{</img>},
\ldots, \\
&\,\boldsymbol{x}_{t_m}, 
\,\texttt{<img>} \,\boldsymbol{x}_{v_m}\, \texttt{</img>},
\mathbf{q}\,].
\end{split}
\end{equation}
Here, $\boldsymbol{x}_{v_k}$ denotes the visual segment of the $k$-th image. Each image is independently encoded at arbitrary resolution, so that the number of visual tokens adapts to its spatial size rather than being constrained to a fixed token budget. This allows different images to preserve visual details at different granularities, which is beneficial for fine-grained comparison and spatially sensitive tasks.

For video inputs, NEO-ov represents the video as a temporally ordered sequence of sampled frames rather than a single global embedding. Specifically, we sample $f$ frames from the raw video and serialize each frame as an image unit associated with a timestamp. Here we further prepend temporal cues to facilitate temporal localization and cross-frame reasoning.
Given sampled frames with timestamps $\tau_1,\ldots,\tau_f$, the video input is written as
\begin{equation}
\small
\begin{split}
\mathbf{X}_{\text{video}} = [\,\mathbf{p}_{\text{global}}, &\,[\tau_1]:
\texttt{<img>} \,\mathbf{x}_{v_1}\, \texttt{</img>}, \,\ldots, \\
&\,[\tau_f]:
\texttt{<img>} \,\mathbf{x}_{v_f}\, \texttt{</img>},
\mathbf{q}\,].
\end{split}
\end{equation}
Here, $\mathbf{p}_{\text{global}}$ denotes a global prefix encoding the video duration, the number of sampled frames, and the sampling rate when available. Temporal information is conveyed jointly with explicit timestamps and frame order within the unified sequence, allowing video understanding to emerge naturally within the same framework as multi-image understanding.

\begin{figure*}[t]
\centering 
\includegraphics[width=0.99\linewidth,trim= 0 0 0 0,clip]{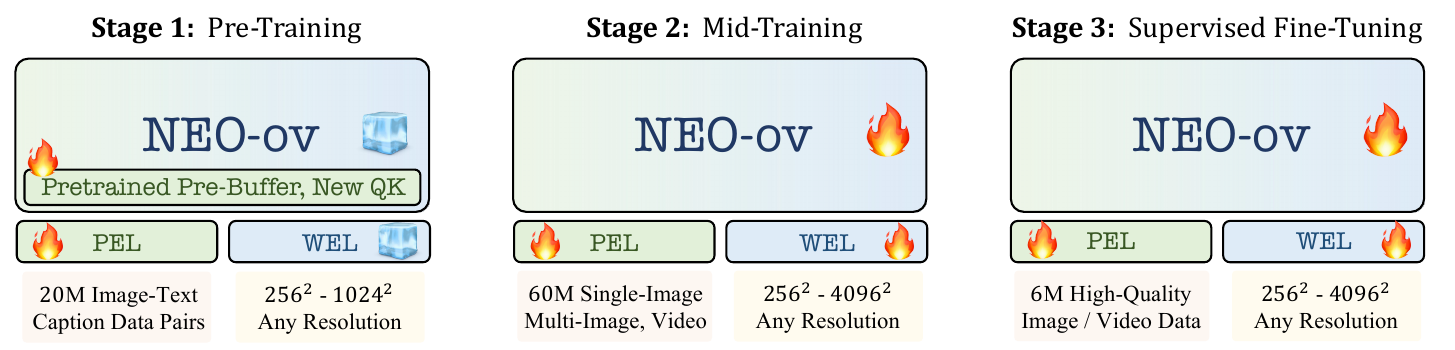} 
\caption{Overview of three-stage training recipe. 
NEO-ov first aligns the Pre-Buffer with the post-LLM using large-scale image-text data while preserving the language abilities of the pretrained LLM. After that, it is optimized with diverse image and video training data to improve spatial-temporal reasoning. Finally, high-quality instruction tuning data further enhances general multimodal understanding, fine-grained perception, and temporal dynamics.}
\label{fig:trainning_recipe}
\end{figure*}

\subsection{Unified Spatial-Temporal Attention}

Compared with single-image modeling, the central challenge in multi-image and video understanding lies not merely in handling longer sequences, but in enabling coherent interactions across multiple visual units within a unified backbone. To address this, we extend native mixed attention from a single visual unit to multiple images and temporally ordered video frames, allowing spatial and temporal dependencies to emerge jointly within the same end-to-end autoregressive framework.

In Figure~\ref{fig:primitive} (2), we treat each image or sampled frame as an independent visual unit. Tokens within the same visual unit attend bidirectionally, while interactions across different visual units remain autoregressive. Let $u_i$ denote the visual unit index of token $i$, where $u_i = 0$ indicates a text token and $u_i > 0$ denotes a visual token from an image or video frame. The attention mask is defined as
\begin{equation}
\mathcal{M}_{ij}=1
\iff
\big(j \le i\big)\ \lor\ \big(u_i=u_j>0\big).
\end{equation}

This design yields two important properties. First, tokens within the same visual unit attend bidirectionally, enabling dense spatial interactions inside each image or frame and allowing rich intra-image structure to be modeled directly. Second, interactions across different visual units remain causal, such that each unit can attend to all preceding text and visual tokens. Unlike modular VLMs, where cross-image or cross-frame reasoning operates on representations already compressed by an external visual encoder, our design allows interactions to emerge directly from patch-level tokens at the earliest layers of the backbone and evolve progressively throughout the network. Consequently, cross-image comparison and temporal reasoning are refined jointly from shallow to deep layers, enabling more precise modeling of fine-grained visual differences and subtle temporal dynamics.

\subsection{Training Procedure}
Our training covers three progressive stages: pre-training, mid-training, and supervised fine-tuning.

\textbf{Pre-Training Stage.}
At this stage, the model develops foundational visual perception while progressively aligning visual representations with the semantic space of the pretrained language backbone. Training is conducted on approximately 20M large-scale image–text pairs collected from diverse web sources, spanning both descriptive captions and OCR-intensive content.
To preserve the linguistic priors of the pretrained LLM and ensure stable multimodal adaptation, optimization is restricted to the patch embedding layers, pre-buffer layers, and newly introduced QK-related parameters. An autoregressive next-token objective aligns visual tokens with the LLM representation space, while pretrained buffer initialization and expanded QK capacity allow visual specialization to emerge without compromising language performance.

\textbf{Mid-Training Stage.}
This stage focuses on scaling spatial-temporal reasoning and enhancing perception over high-resolution visual content. Training continues on nearly 60M multimodal samples, covering resolutions from $256^2$ to $4096^2$ and videos of up to 128 frames.
At this stage, all model layers are jointly optimized to strengthen cross-modal interaction and contextual coherence across both pixel-world and pixel-pixel relations. The context length is progressively extended from 16K to 36K tokens, enabling more effective modeling of high-resolution inputs and long video sequences. To support diverse application scenarios, we adopt a unified mixture of text-only, image-text, multi-image, and video-text data with an approximate ratio of 2:4:1:1, improving optimization stability and generalization across heterogeneous tasks.

\textbf{Supervised Fine-Tuning Stage.}
In this stage, the model is refined using high-quality instruction-tuning data, including approximately 4M single-image, 1M multi-image, and 1M video samples, to enhance multimodal understanding and cross-frame reasoning. The training corpus covers visual question answering, OCR understanding, fine-grained perception, temporal reasoning, mathematical analysis, and complex dialogue.
The entire model is optimized end-to-end under next-token prediction objectives, further strengthening fine-grained perception, long-context reasoning, and temporal dynamics modeling. Combined with multi-resolution training up to $4096^2$ and videos of up to 128 frames, this stage equips the model with strong generalization across a wide range of real-world multimodal visual understanding tasks.



\begin{table*}[t!]
\centering
\renewcommand{\arraystretch}{1.15}
\resizebox{\linewidth}{!}{
\begin{tabular}{lccccccccccc}
\toprule
\multirow{2}{*}{\textbf{Model}}
&\multicolumn{6}{c}{\textbf{General VQA Understanding}}
&\multicolumn{5}{c}{\textbf{OCR Recognization}} \\
\cmidrule(lr){2-7}\cmidrule(lr){8-12}
&\textbf{MMMU} &\textbf{MMB} &\textbf{RWQA} &\textbf{MMStar} &\textbf{SEED-I} &\textbf{HallB} &\textbf{AI2D} &\textbf{DocVQA} &\textbf{ChartQA} &\textbf{TextVQA} &\textbf{OCRBench}
\\
\midrule
\rowcolor{tablerowcolor1} \multicolumn{12}{l}{$\blacktriangledown$ \emph{Modular Vision-Language Models (Instruct-2B)}} \\

Qwen2-VL &41.1 &74.9 &62.6 &48.0 &-- &41.7 &74.7 &90.1 &73.5 &\textbf{79.7} &80.9 \\

InternVL3 &48.6 &\textbf{81.1} &\textbf{64.3} &60.7 &-- &42.5 &78.7 &88.3 &80.2 &77.0 &83.5 \\

InternVL3.5 &53.0 &78.2 &62.0 &\textbf{62.7} &\textbf{75.3} &48.6 &\textbf{78.8} &89.4 &\textbf{80.7} &76.5 &83.6 \\

Qwen3-VL & \textbf{53.4} &78.4 &63.9 &58.3 &-- &\textbf{51.4} &76.9 &\textbf{93.3} &79.1 &-- &\textbf{85.8} \\

\midrule
\rowcolor{tablerowcolor} \multicolumn{12}{l}{$\blacktriangledown$ \emph{Native Vision-Language Models (Instruct-2B)}} \\

Mono-VL &33.7 &65.5 &-- &-- &67.4 &34.8
&68.6 &80.0 &73.7 &72.6 &76.7 \\

Mono-VL1.5 &39.1 &64.0 &-- &-- &66.9 &32.5
&67.4 &81.7 &72.2 &73.7 &80.1 \\

HoVLE &32.2 &73.3 &-- &-- &70.9 &38.4
&73.0 &86.1 &78.6 &70.9 &74.0 \\

OneCAT &39.0 &72.4 &-- &-- &70.9 &--
&72.4 &87.1 &76.2 &67.0 &-- \\

NEO &48.6 &76.0 &63.1 &54.2 &74.2 &43.1
&80.1 &89.9 &81.2 &74.0 &77.1 \\

\rowcolor{gray!18} \textbf{NEO-ov} &\textbf{54.7} &\textbf{80.0} &\textbf{64.4} &\textbf{58.6} &\textbf{76.2} &\textbf{54.5} &\textbf{81.4} &\textbf{91.2} &\textbf{83.1}  &\textbf{77.3} &\textbf{81.2} \\

\midrule
\rowcolor{tablerowcolor1} \multicolumn{12}{l}{$\blacktriangledown$ \emph{Modular Vision-Language Models (Instruct-8B)}} \\

Qwen2.5-VL &55.0 &83.5 &68.5 &63.9 &-- &52.9 &83.9 &95.7 &87.3 &\textbf{84.9} &86.4 \\

InternVL3 &62.7 &83.4 &70.8 &68.2 &-- &49.9
&85.2 &92.7 &86.6 &80.2 &88.0 \\

InternVL3.5 &68.1 &82.7 &67.5 &69.3 &\textbf{77.1} &54.5 &84.0 &92.3 &86.7 &78.2 &84.0 \\

Qwen3-VL &\textbf{69.6} &\textbf{84.5} &\textbf{71.5} &\textbf{70.9} &-- &\textbf{61.1} &\textbf{85.7} &\textbf{96.1} &\textbf{89.6} &-- &\textbf{89.6} \\

\midrule

\rowcolor{tablerowcolor} \multicolumn{12}{l}{$\blacktriangledown$ \emph{Native Vision-Language Models (Instruct-8B)}} \\

Fuyu &27.9 & 10.7 & 43.7 & -- & 59.3 & --  & 64.5 & -- & -- & -- & 36.6 \\

EVE &32.6& 52.3 & -- & -- & 64.6 & 26.4 & 61.0 & 53.0 & 59.1 & 56.8 & 39.8 \\

SOLO &-- & 67.7 & 44.7 & -- & 64.4 & -- & 61.4 & -- & -- & -- & 12.6 \\

EVEv2 &39.3 & 66.3 &62.4 & -- & 71.4  & -- & 74.8 & -- & 73.9 & 71.1 & 70.2 \\

BREEN &42.7 & 71.4 & -- & 51.2 & -- & 37.0 & 76.4 & -- & -- & 65.7 & -- \\

VoRA &32.0 & 61.3 & 60.1 & -- & 68.9 & -- & 61.1 & -- & -- & 58.7 & -- \\

SAIL &-- & 70.1 & 63.9 & 53.1 & 72.9 & 54.2 & 76.7 & -- & --  & 77.1 & 78.3 \\

NEO &54.6 &82.1 &67.3 &62.4 &76.3 &46.4 &83.1 &88.6 &82.1 &75.0 &77.7 \\

\rowcolor{gray!18} 
\textbf{NEO-ov} &\textbf{68.1} &\textbf{85.1} &\textbf{67.8} &\textbf{67.3} &\textbf{76.6} &\textbf{59.8} &\textbf{85.4} &\textbf{91.9} &\textbf{86.2} &\textbf{78.5} &\textbf{81.6}\\
\bottomrule
\end{tabular}
}
\caption{Comparison with existing popular VLMs on general VQA and OCR benchmarks.}
\setlength\tabcolsep{2pt}
\label{tab:results_image}
\end{table*}

\section{Experiment}

\subsection{Implementation Details}

The NEO-ov model is trained on sixteen 8-GPU nodes, each equipped with 80 GB GPUs. Here we use the AdamW optimizer~\citep{Training:AdamW} with cosine learning-rate decay and a warm-up ratio of 0.01. The peak learning rates for the three training stages are set to $2 \times 10^{-4}$, $5 \times 10^{-5}$, and $5 \times 10^{-5}$, respectively.
We use Qwen3-1.7B and Qwen3-8B~\citep{TransF:Qwen3} as the language backbones. The pre-buffer module consists of 12 layers for NEO-ov (2B) and 6 layers for NEO-ov (9B). The native RoPE base frequencies, $\theta_T$, $\theta_H$, and $\theta_W$, are fixed at $1 \times 10^{6}$, $1 \times 10^{4}$, and $1 \times 10^{4}$.

\subsection{Main Results}

We evaluate NEO-ov using VLMEvalKit~\cite{VLM:VLMEvalKit} on three domains: image understanding, video understanding, and spatial intelligence.

\textbf{Image Understanding.}
We test NEO-ov on general visual perception and reasoning benchmarks such as MMMU~\cite{Datasets:MMMU}, MMBench-EN (MMB)~\cite{Datasets:MMBench}, RealWorldQA (RWQA)~\cite{Datasets:Realworldqa}, MMStar~\cite{Datasets:MMStar}, and SEEDBench-IMG (SEED-I)~\cite{Datasets:Seed-bench}; document, diagram, chart, and text understanding benchmarks including AI2D~\cite{Datasets:AI2D}, DocVQA~\cite{Datasets:DocVQA}, ChartQA~\cite{Datasets:ChartQA}, InfoVQA~\cite{Datasets:InfoVQA}, TextVQA~\cite{Datasets:TextVQA}, and OCRBench~\cite{Datasets:OCRBench}; hallucination task on HallusionBench (HallB)~\cite{Datasets:Hallusionbench}.

\begin{table*}[t]
\centering
\setlength\tabcolsep{6pt}
\renewcommand{\arraystretch}{1.15}

\resizebox{\linewidth}{!}{
\begin{tabular}{lcccccccc}
\toprule

\multirow{2}{*}{\textbf{Model}}
& \multicolumn{2}{c}{\textbf{Multi-Image}}
& \multicolumn{6}{c}{\textbf{Video Understanding}}
\\
\cmidrule(lr){2-3}
\cmidrule(lr){4-9}
& \textbf{BLINK}
& \textbf{MUIRBENCH}
& \textbf{VideoMME}
& \textbf{MVBench}
& \textbf{LVBench}
& \textbf{MLVU}
& \textbf{LongVideoBench}
& \textbf{VideoMMMU}
\\
\midrule
\rowcolor{tablerowcolor1}
\multicolumn{9}{l}{$\blacktriangledown$ \emph{Modular Vision-Language Models (Instruct-2B)}} \\
VideoLLaMA3
&44.2 &-- 
&59.6 &65.5 &41.6 &65.4 &57.1 &-- \\
InternVL3.5
&51.3 &44.0
&58.4 &\textbf{65.9} &37.6 &64.4 &\textbf{57.4} &\textbf{42.7} \\
Qwen3-VL
&\textbf{53.8} &\textbf{47.4}
&\textbf{61.9} &61.7 &\textbf{47.4} &\textbf{68.3} &55.6 &41.9 \\
\midrule
\rowcolor{tablerowcolor} \multicolumn{9}{l}{$\blacktriangledown$ \emph{Native Vision-Language Models (Instruct-2B)}} \\
ELVA
&-- &--
&41.8 &43.5 &-- &47.6 &-- &-- \\
\rowcolor{gray!18}
\textbf{NEO-ov}
&\textbf{53.9} &\textbf{56.8}
&\textbf{60.4} &\textbf{65.7} &\textbf{43.3} &\textbf{64.8} &\textbf{56.8} &\textbf{42.3} \\
\midrule
\rowcolor{tablerowcolor1}
\multicolumn{9}{l}{$\blacktriangledown$ \emph{Modular Vision-Language Models (Instruct-8B)}} \\
LLaVA-Video
&-- &--
&63.3 &58.6 &44.2 &70.8 &58.2 &-- \\
VideoLLaMA3
&56.7 &-- 
&66.2 &69.7 &45.3 &73.0 &59.8 &-- \\
InternVL3.5
&59.5 &55.8
&66.0 &\textbf{72.1} &45.9 &70.2 &62.1 &54.9 \\
Qwen3-VL
&\textbf{69.1} &\textbf{64.4}
&\textbf{71.4} &68.7 &\textbf{58.0} &\textbf{78.1} &\textbf{63.6} &\textbf{65.3} \\
\midrule
\rowcolor{tablerowcolor} \multicolumn{9}{l}{$\blacktriangledown$ \emph{Native Vision-Language Models (Instruct-8B)}} \\
Fuyu
&-- &--
&28.7 &31.6 &-- &31.1 &-- &-- \\
EVE
&-- &--
&29.3 &34.9 &-- &36.8 &-- &-- \\
ELVA
&-- &--
&47.1 &51.2 &-- &51.8 &-- &-- \\
\rowcolor{gray!18}
\textbf{NEO-ov}
&\textbf{62.8} &\textbf{58.2}
&\textbf{67.4} &\textbf{70.7} &\textbf{46.4} &\textbf{69.3} &\textbf{63.5} &\textbf{51.6} \\

\bottomrule
\end{tabular}
}
\caption{Comparison with existing popular VLMs on multi-image and video benchmarks.}
\label{tab:results_video}
\end{table*}

\begin{table*}[t!]
\centering
\setlength\tabcolsep{6pt}
\renewcommand{\arraystretch}{1.15}
\resizebox{\linewidth}{!}{
\begin{tabular}{lccccccccc}
\toprule
\textbf{Model}
& \textbf{VSI-Bench}
& \textbf{MMSI}
& \textbf{Mindcube}
& \textbf{ViewSpatial}
& \textbf{SITE}
& \textbf{3DSR}
& \textbf{EmbSpatial}
& \textbf{SPAR}
& \textbf{Omni-Spatial}
\\
\midrule

\rowcolor{tablerowcolor1} \multicolumn{10}{l}{$\blacktriangledown$ \emph{Spatial-specialist Models (Instruct-2B)}} \\
Cambrian-S (3B)
&56.1 &27.0 &38.4 &41.0 &31.0
&41.4 &\textbf{63.5} &33.0 &\textbf{41.9} \\
Sensenova-SI
&\textbf{63.7} &\textbf{34.2} &\textbf{41.8} &\textbf{52.7} &\textbf{36.8} &\textbf{50.5} &62.8 &\textbf{38.0} &26.4 \\
\midrule
\rowcolor{tablerowcolor} \multicolumn{10}{l}{$\blacktriangledown$ \emph{General-purpose Models (Instruct-2B)}} \\
InternVL3.5 &53.8 &25.6 &42.1 &37.9 &34.8 &31.4 &61.5 &32.4 &\textbf{44.4} \\
Qwen3-VL &53.9 &27.8 &34.2 &36.7 &35.8 &47.6 &\textbf{69.2} &34.1 &36.3 \\
\rowcolor{gray!18}
\textbf{NEO-ov} &\textbf{58.4} &\textbf{33.6} &\textbf{77.2} &\textbf{52.8} &\textbf{38.4} &\textbf{52.9} &63.8 &\textbf{41.2} &43.1 \\
\midrule
\rowcolor{tablerowcolor1} \multicolumn{10}{l}{$\blacktriangledown$ \emph{Spatial-specialist Models (Instruct-8B)}} \\
Cambrian-S
&67.5 &25.8 &39.6 &40.9 &33.0 
&45.0 &72.8 &37.9 &\textbf{41.9} \\
Sensenova-SI
&68.8 &\textbf{43.3} &\textbf{85.7} &\textbf{54.7} &47.7 
&\textbf{55.5} &72.0 &45.8 &33.0 \\
GeoThinker
&\textbf{72.6} &30.9 &83.0 &45.9 &\textbf{55.9}
&51.9 &\textbf{78.8} &\textbf{68.2} &40.1 \\
\midrule
\rowcolor{tablerowcolor} \multicolumn{10}{l}{$\blacktriangledown$ \emph{General-purpose Models (Instruct-8B)}} \\
InternVL3.5 &56.3 &29.1 &40.4 &40.0 &\textbf{54.4} &35.3 &75.7 &38.2 &\textbf{47.8}  \\
Qwen3-VL &59.4 &31.2 &29.6 &41.9 &45.4 &52.9 &77.8 &40.3 &47.0  \\
\rowcolor{gray!18}
\textbf{NEO-ov} &\textbf{64.8} &\textbf{41.3} &\textbf{90.0} &\textbf{55.2} &54.3 &\textbf{61.7} &\textbf{78.8} &\textbf{48.8} &45.0 \\
\bottomrule
\end{tabular}
}
\caption{Comparison with existing popular VLMs on spatial intelligence benchmarks.}
\label{tab:results_spatial}
\end{table*}

\textit{Comparison with Native VLMs.}
As shown in Table~\ref{tab:results_image}, 
NEO-ov establishes a new performance frontier for native VLMs at both 2B and 8B scales, consistently surpassing prior native architectures including NEO~\cite{VLM:NEO}, EVE series~\cite{VLM:EVE,VLM:EVEv2}, Mono-InternVL series~\cite{VLM:Mono-InternVL,VLM:Mono-InternVL-1.5}, OneCAT~\cite{VLM:OneCAT}, Emu3~\cite{VLM:Emu3}, and SAIL~\cite{VLM:SAIL}. The gains are particularly pronounced on reasoning-intensive and hallucination-sensitive benchmarks such as MMMU, HallB, and InfoVQA, demonstrating that native end-to-end modeling can unlock strong visual reasoning and representation learning even without external visual encoders. It further underscores the scalability and emerging competitiveness of the native one-vision paradigm.

\begin{figure*}[t]
    \begin{minipage}[t]{0.47\textwidth}
    \centering
    \includegraphics[width=\linewidth,trim= 0 0 0 0,clip]{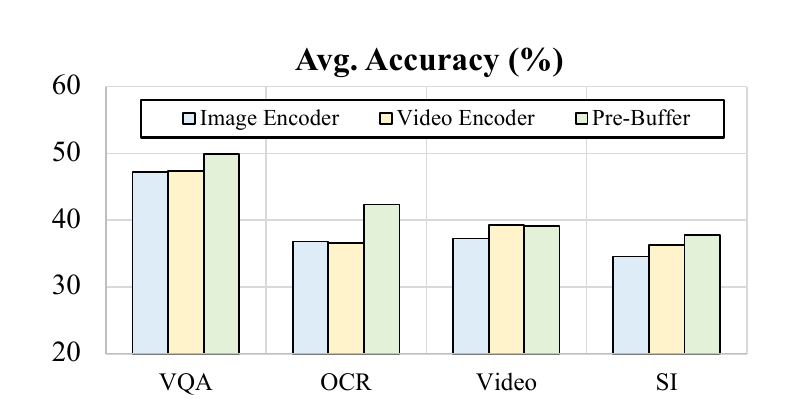}
    \vspace{-1.5em}
    \caption{Pre-Buffer vs. VEs on diverse tasks.}
    \label{fig:compare_encoder}
    \hfill
    \end{minipage}
    \begin{minipage}[t]{0.28\textwidth}
    \centering
    \includegraphics[width=\linewidth,trim= 0 0 0 0,clip]{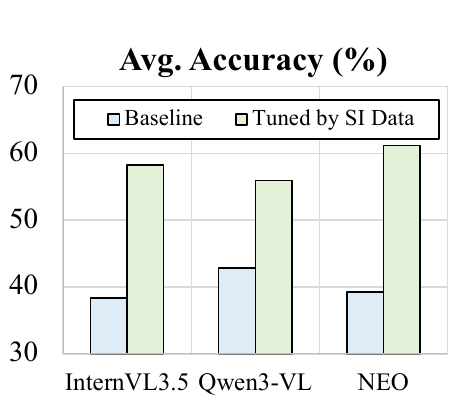}
    \vspace{-1.5em}
    \caption{Finetuned on SI data.}
    \label{fig:compare_si}
    \end{minipage}
    \hfill
    \begin{minipage}[t]{0.235\textwidth}
    \centering    
    \includegraphics[width=\linewidth,trim= 0 0 0 0,clip]{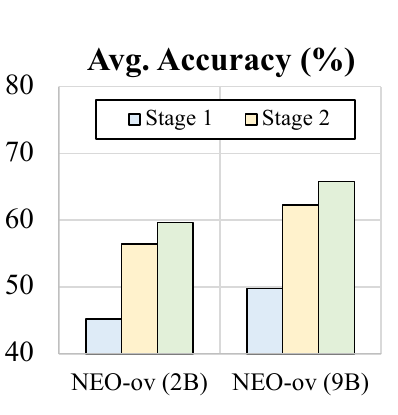} 
    \vspace{-1.5em}
    \caption{Three stages.}
    \label{fig:stage_results}
    \end{minipage}
\end{figure*}

\textit{Comparison with Modular VLMs.}
Beyond native models, NEO-ov also demonstrates strong competitiveness against leading modular VLMs such as InternVL3.5~\cite{VLM:InternVL-3.5} and Qwen3-VL~\cite{VLM:Qwen3VL}. Despite operating without pretrained visual encoders, NEO-ov matches or surpasses its modular counterpart~\cite{VLM:InternVL-3.5} on several reasoning and perception benchmarks, particularly in complex reasoning and hallucination suppression. 
While OCR-intensive tasks remain challenging, native architectures are rapidly closing the gap with modular systems across diverse image understanding benchmarks. Overall, these findings further validate the competitiveness and scalability of fully native multimodal modeling.

\textbf{Multi-Image and Video Understanding.}
Compared with prior native VLMs such as Fuyu~\cite{VLM:Fuyu-8b}, EVE~\cite{VLM:EVE}, and ELVA~\cite{VLM:ELVA} in Table~\ref{tab:results_video}, NEO-ov achieves substantial gains on VideoMME~\cite{fu2025videomme}, MVBench~\cite{li2024mvbench}, and MLVU~\cite{zhou2025mlvu}, highlighting its strong temporal reasoning and long-context visual understanding capabilities at both 2B and 8B scales.
It also remains highly competitive with several modular VLMs, including VideoLLaMA3~\cite{VLM:VideoLLaMA3} and InternVL3.5~\cite{VLM:InternVL-3.5} on BLINK~\cite{fu2024blink}, MUIRBENCH~\cite{wang2025muirbench}, LVBench~\cite{wang2025lvbench}, LongVideoBench~\cite{wu2024longvideobench}, and VideoMMMU~\cite{hu2025videommmu}. These results indicate that a unified native backbone can naturally support cross-image reasoning and temporal association within a single autoregressive framework.

\textbf{Spatial Intelligence.}
In Table~\ref{tab:results_spatial}, NEO-ov displays strong spatial intelligence across geometric reasoning, spatial perception, and embodied understanding benchmarks. Compared with spatial-specialist models such as Cambrian-S~\cite{yang2025cambrian-s}, Sensenova-SI~\cite{cai2025sensenova-si}, and GeoThinker~\cite{li2026geothinker}, NEO-ov, as a general-purpose native VLM, achieves comparable or even better performance at both 2B and 8B scales. In particular, NEO-ov shows clear advantages over other general VLMs on VSI-Bench~\cite{yang2025vsi-bench}, MMSI~\cite{yang2025mmsi}, Mindcube-tiny (Mindcube)~\cite{wang2025mindcube}, ViewSpatial~\cite{li2025viewspatial}, SITE~\cite{wang2025site}, 3DSR~\cite{ma20253dsrbench}, EmbSpatial~\cite{du2024embspatial}, SPAR~\cite{zhang2026spar}, and Omni-Spatial (manual CoT)~\cite{jia2025omnispatial}, highlighting its ability to capture fine-grained spatial and geometric representations.


\subsection{Ablation Studies}
\textbf{Native Attention vs. Encoder-based Attention.}
Figure~\ref{fig:compare_encoder} compares the Pre-Buffer mechanism with conventional visual encoders across diverse tasks, including general VQA, OCR, video understanding (Video), and spatial intelligence (SI). Both architectures are randomly initialized for fair comparison. In image encoders, attention is restricted to bidirectional interactions among visual tokens within the same image, while video encoders further extend such interactions across frames.
We can observe that Pre-Buffer consistently achieves competitive or superior performance across all benchmarks, especially on OCR and SI tasks, where fine-grained visual structure and long-range spatial dependencies are especially critical. These gains suggest that preserving richer intermediate visual context through native pixel-pixel and pixel-word interactions is more effective than relying solely on compressed image- or video-level representations. Moreover, the consistent performance across VQA, OCR, Video, and SI benchmarks highlights the strong generalization capability of native architectures under diverse multimodal scenarios.

\textbf{Deep Interactions Benefit Spatial Intelligence.}
Figure~\ref{fig:compare_si} highlights a clear advantage of native architectures on spatial intelligence tasks. Although all models benefit from additional SI supervision, NEO shows substantially larger gains than encoder-based models such as InternVL3.5 and Qwen3-VL. We attribute this to the native interaction pattern of NEO, where pixel-pixel and pixel-word interactions emerge directly in shallow layers of the unified backbone, enabling richer spatial and cross-modal representations from the early fusion.

\textbf{Performance Improvements across Stages.}
Figure~\ref{fig:stage_results} illustrates performance evolution across all single-image, multi-image, video, and spatial intelligence benchmarks. Performance improves consistently from Stage 1 to Stage 2 for both the 2B and 9B variants of NEO-ov, with especially pronounced gains at smaller scales. These results suggest that progressive training effectively strengthens general visual understanding and leads to more robust multimodal capabilities across diverse tasks.

\section{Conclusion}

In this paper, we launch NEO-ov, a fully native vision–language foundation model that unifies single-image understanding, multi-image reasoning, video comprehension, and spatial intelligence within a single monolithic backbone. Unlike conventional modular VLMs, NEO-ov learns visual perception, temporal dynamics, and cross-modal correspondence directly from raw inputs through end-to-end training, without relying on external visual encoders.
Extensive experiments demonstrate that NEO-ov achieves competitive performance against strong encoder-based counterparts while showing clear advantages in fine-grained perception and spatial reasoning. Beyond empirical results, our findings suggest that unified native architectures provide a promising path toward scalable and general-purpose one-vision foundation models.

\section{Limitations}

Despite the strong empirical performance of NEO-ov, several challenges remain open for future exploration. First, although NEO-ov substantially advances native vision-language modeling, a gap still exists between NEO-ov and top-tier modular systems such as Qwen3-VL on certain single-image and video understanding benchmarks. We believe this gap is largely attributable to the current scale and quality of multimodal training data, particularly for complex reasoning, temporal perception, and fine-grained visual-text alignment.

Second, OCR-intensive and document-centric tasks remain relatively underexplored for native architectures. Unlike modular VLMs that benefit from specialized visual encoders and extensive OCR-oriented pretraining, NEO-ov currently lacks sufficiently diverse and high-quality supervision for documents, charts, and dense text perception. We expect that improving OCR-related data scales and quality will further strengthen them.

Finally, while NEO-ov already shows promising capabilities in multi-image reasoning, video understanding, and spatial intelligence, the broader potential of native multimodal modeling remains far from fully explored. Further scaling in model capacity, multimodal data diversity, and long-context training may unlock substantially stronger multimodal reasoning and perception capabilities.

\section{Ethical Considerations}
All resources are drawn from open-access datasets with explicitly defined usage policies. Our work seeks to advance multimodal learning capabilities without introducing ethical or safety concerns beyond those already associated with existing models. Nevertheless, risks such as dataset biases and potential misuse cannot be entirely ruled out. We emphasize the importance of careful data curation, responsible deployment, and transparent reporting as essential practices to mitigate these challenges. 

During manuscript preparation, large language models were used solely as writing assistants. 
They helped to check grammar, refine sentence structure, and provide style alternatives. 
All content related to methodology, experiments, and conclusions was developed entirely by the authors. 
LLM outputs were reviewed critically, and only human-verified edits were incorporated into the final text. 


\bibliography{custom}

\end{document}